\pdfoutput=1

\documentclass[11pt]{article}

\usepackage{emnlp2022}
\usepackage{CJKutf8}
\usepackage{times}
\usepackage{latexsym}
\usepackage{listings}
\usepackage{graphicx}
\usepackage{float}
\usepackage{subfigure}
\usepackage{xcolor}
\usepackage{amsmath}
\usepackage{amsfonts}
\usepackage{algorithm}
\usepackage[noend]{algpseudocode}
\usepackage{multirow}
\usepackage{booktabs,hhline}
\usepackage{float}
\usepackage{hyperref}
\hypersetup{
    colorlinks=true,
    linkcolor=blue,
    filecolor=magenta,
    urlcolor=darkblue,
    pdfpagemode=FullScreen,
}
\usepackage{makecell}

\usepackage[T1]{fontenc}

\usepackage[utf8]{inputenc}

\usepackage{microtype}

\title{Doc2Bot: Accessing Heterogeneous Documents via Conversational Bots}
\author{Haomin Fu\textsuperscript{1,2}\footnotemark[1], Yeqin Zhang\textsuperscript{1,2}\footnotemark[1], Haiyang Yu\textsuperscript{2}, Jian Sun\textsuperscript{2}, Fei Huang\textsuperscript{2}, Luo Si\textsuperscript{2} \\ \textbf{Yongbin Li\textsuperscript{2}\footnotemark[2] \and Cam-Tu Nguyen\textsuperscript{1}\footnotemark[2]}\\
  \textsuperscript{1}State Key Laboratory for Novel Software Technology, Nanjing University, China \\
  \textsuperscript{2}Alibaba Group \\
  \{haominfu, zhangyeqin\}@smail.nju.edu.cn \\
  \{yifei.yhy, jian.sun, f.huang, luo.si\}@alibaba-inc.com \\
  shuide.lyb@alibaba-inc.com, ncamtu@nju.edu.cn
  }

\newcommand{\dsname}{Doc2Bot}

\begin{document}
\maketitle
\renewcommand{\thefootnote}{\fnsymbol{footnote}}
\footnotetext[1]{Equal contribution.}
\footnotetext[2]{Corresponding authors.}
\renewcommand{\thefootnote}{\arabic{footnote}}
\begin{abstract}

%

This paper introduces {\dsname}, a novel dataset for building machines that help users seek information via conversations. This is of particular interest for companies and organizations that own a large number of manuals or instruction books. Despite its potential,  the nature of our task poses several challenges: (1) documents contain various structures that hinder the ability of machines to comprehend, and (2) user information needs are often underspecified. Compared to prior datasets that either focus on a single structural type or overlook the role of questioning to uncover user needs, the {\dsname} dataset is developed to target such challenges systematically. Our dataset contains over 100,000 turns based on Chinese documents from five domains, larger than any prior document-grounded dialog dataset for information seeking. We propose three tasks in {\dsname}: (1) dialog state tracking to track user intentions, (2) dialog policy learning to plan system actions and contents, and (3) response generation which generates responses based on the outputs of the dialog policy. Baseline methods based on the latest deep learning models are presented, indicating that our proposed tasks are challenging and worthy of further research. 
\end{abstract}

\section{Introduction}

The last decade has witnessed a dramatic change in how humans interact with information retrieval systems. Although traditional search engines still play an important role in our daily life, the wide adoption of smart devices with small screens requires systems to answer user requests more concisely. Early attempts focus on answering independent questions \cite{rajpurkar2016squad}, whereas recent studies pay attention to handling interconnected questions via conversations around a single passage \cite{pasupat-liang2015compositional, chen2020hybridqa} or documents \cite{feng2020doc2dial, feng2021multidoc2dial}. Yet, the nature of heterogeneous documents and our conversational setting pose challenges that require further attention. We, therefore, develop {\dsname}\footnote{\href{https://github.com/Doc2Bot/Doc2Bot}{https://github.com/Doc2Bot/Doc2Bot}} with these considerations in mind.

\begin{figure*}[htbp]
\centering
\includegraphics[width=0.98\textwidth]{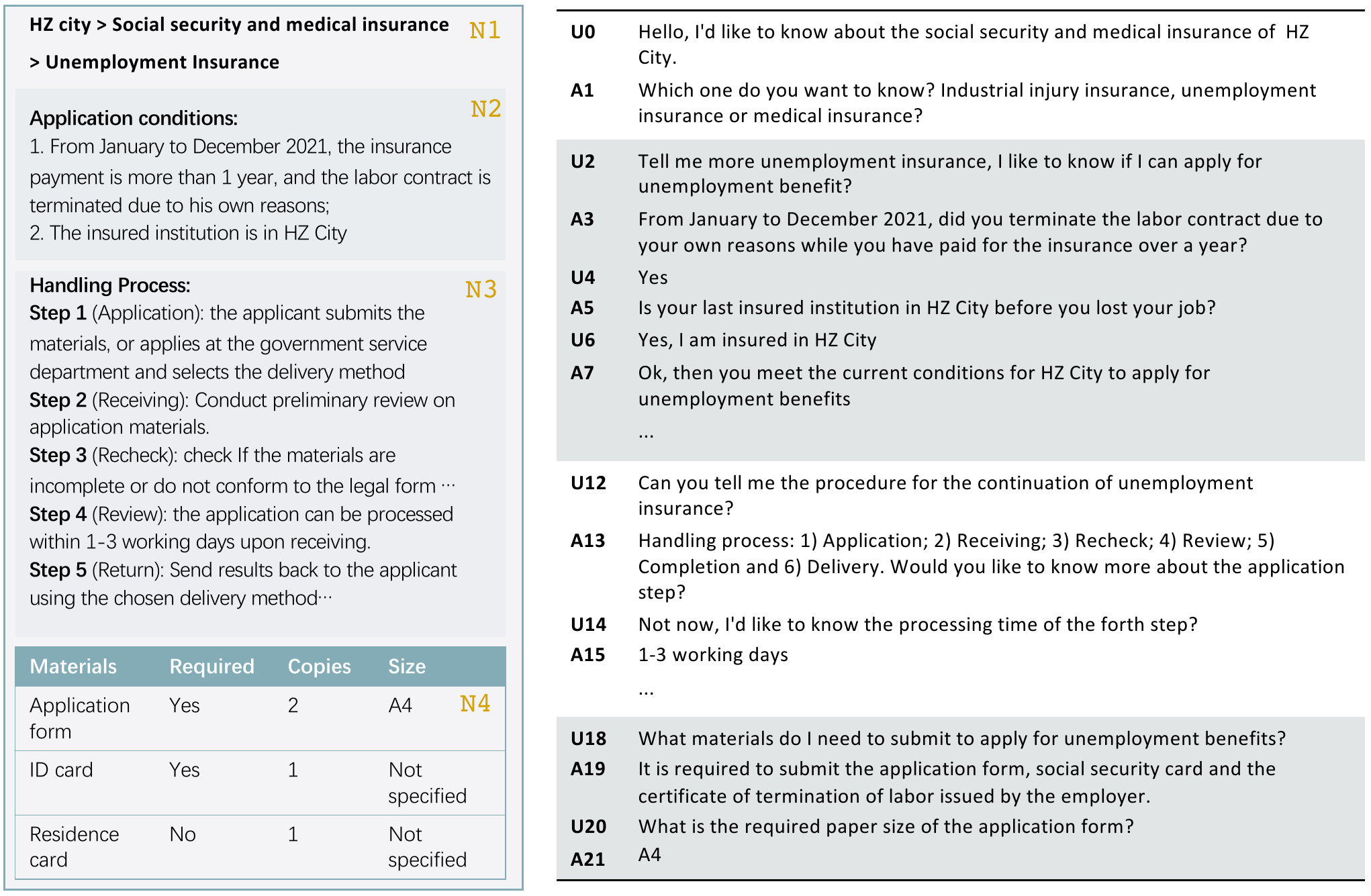}
\caption{An example dialog (right) grounded on a document (left) with heterogeneous structures. From the top, the dialog contains 4 segments \texttt{S1-4} grounded on 4 corresponding document segments \texttt{N1-4}. Here \texttt{U} and \texttt{A} stand for user and agent, respectively.}
\label{fig:example}
\end{figure*}

The first concerns the nature of heterogeneous documents, which often contain different types of structures such as tables and sequences. To answer questions regarding such structural types, systems need to acquire various skills. Figure \ref{fig:example} shows a conversation between a user and an agent, where the agent has access to a collection of documents. In this conversation, every utterance except the first one depends on the conversation history and the grounded documents. The document contains diverse structures including conditions (\texttt{N2}), sequences (\texttt{N3}), and tables (\texttt{N4}). Each type of structure poses different challenges for the agent. For example, the agent needs to understand the rule text in \texttt{N2} to ask questions such as \texttt{A3} and \texttt{A5}. The agent also needs to be able to summarize the procedure in \texttt{N3} to produce the answer (\texttt{A13}) or jump to a specific step of the procedure if being asked (\texttt{A15}). In addition, the agent needs to be capable of summarizing the tables (\texttt{A19}) or answering questions with row and column constraints (\texttt{A21}). Note that there have been previous attempts \cite{pasupat-liang2015compositional, saeidi2018interpretation, chen2020hybridqa,zhang2022layout} to address different structures in isolation. However, little research has been done on handling multiple types of structures in the same document context.

The second concerns the nature of our conversational setting, which is to help users seek information from documents. Since user information needs are often under-specified, it is desirable for systems to ask clarifying questions. This resembles the task of relevance feedback and query refinement in traditional information retrieval. However, in {\dsname}, system feedback is in the form of natural questions, and thus more user-friendly. For example, in Figure \ref{fig:example}, \texttt{A1} is a kind of multiple-choice question that the agent asks to narrow down the search for the answer. In contrast, \texttt{A3} and \texttt{A5} are to verify user situations to answer questions regarding condition/solution structure. Although learning to construct questions from a single passage has been studied in Machine Reading Comprehension \cite{saeidi2018interpretation,guo2021abg}, such finer-grained questions are required only when the passage containing the answer has been found. For document-grounded dialog systems (DGDS), the agent needs to have the skills to narrow down the search (\texttt{A1}) as well as to ask finer questions such as \texttt{A3} and \texttt{A5}.

Towards such goals, there are several challenges that we need to address. First, documents come in different formats, and thus the process of constructing our dataset is more difficult than those from single passages with homogeneous structures. The difference in formats also hinders the ability of machines to learn common patterns. Second, like human-human conversations, it is desirable to have samples of human-system conversations that are natural, and coherent while being diverse for the machine learning purpose. We target such challenges systematically and make the following contributions:

\begin{itemize}
\item We present a unified representation for heterogeneous structures, which not only facilitates our data collection process but also helps systems to learn patterns across documents.  
\item We propose an agenda-based dialog collection protocol that controls the diversity and coherence of dialogues by design. The protocol also encourages crowd-collaborators to introduce ambiguities to conversations.
\item We introduce a new dataset {\dsname} which is larger in scale compared to recent datasets for DGDS \cite{feng2020doc2dial, feng2021multidoc2dial} while introducing new challenges such as a new language (Chinese), richer relations (e.g, sections, conditions, tables, sequences) and new tasks (e.g. dialog policy learning).
\item We evaluate our proposed tasks with the latest machine learning methods. The experiments show that our tasks are still challenging, which suggests room for further research. 
\end{itemize}

\section{Related Works}
Our work is most closely related to the document-grounded dialog systems (DGDS) in the literature. Based on the conversation objective, we can roughly categorize the related tasks into chitchat, comprehension, or information seeking.


Document-grounded chitchat datasets such as WoW \cite{dinan2019wizard}, Holl-E \cite{moghe2018towards}, CMU-DoG \cite{zhou2018dataset} aim to enhance early chitchat systems by using information from grounded textual passages for answer generation. The goal is similar to an open chitchat system as the dialog agent tries to keep users engaged in long, informative, and interactive conversations. This is different from our setting because users of our system often have clear goals (information needs), and the dialog agent needs to provide users with accurate information as soon as possible.

For document-grounded “comprehension” such as CoQA \cite{reddy2019coqa}, Abg-CoQA \cite{guo2021abg} and ShARC \cite{saeidi2018interpretation}, the agent is given a textual paragraph and needs to answer users’ questions about the paragraph. This setting is similar to Machine Reading Comprehension (MRC). However, the difference is that questions in MRC may not form a coherent dialog. Noticeably, several question strategies have been targeted in Abg-CoQA and ShARC. For example, in Abg-CoQA, systems can ask clarifying questions to resolve different types of ambiguities. In ShARC, the authors created conversations where the system can learn to ask “yes/no” questions to understand users’ information and provide appropriate answers. The questioning strategy in ShARC is designed based on text rules that define the relationship between “conditions” and “solutions” exhibited in the given paragraph. Although we also address question strategies, our tasks are more challenging because we focus on multiple documents.

The third type of DGDS \cite{penha2019introducing, feng2020doc2dial, feng2021multidoc2dial} is closest to our setting where the agent needs to provide answers to information seekers in the shortest possible time. Mantis \cite{penha2019introducing} was collected from online forums, and the grounded documents are not given in advance. As a result, Mantis does not come with a detailed annotation which is needed to study the capability of the agents to understand documents. In contrast, given a set of documents, Doc2dial \cite{feng2020doc2dial} and Multidoc2dial \cite{feng2021multidoc2dial} were collected in 2 stages: 1) dialog flows are first generated by labeling and linking paragraphs, 2) crowdsourcers then write conversations based on the suggested flows. Note that Multidoc2dial was built by rearranging dialogues from doc2dial so that one conversation can contain information from multiple documents. Although we follow similar steps for constructing the dataset, our dialog flow generation is essentially different, which addresses the coherence of the generated dialogues, and the multi-document grounding issue by design. In addition, our dataset exceeds Doc2dial and Multidoc2dial in scale, while also highlighting new challenges such as under-specified user requests. 


\section{Dataset Collection}


This section details the process of collecting \dsname, including 4 stages: 1) \textbf{document collection} which selects targeted domains and documents; 2) \textbf{document graph construction} which unifies heterogeneous structures from multiple domains to build document graphs; 3) \textbf{dialog flow generation} that simulates the agenda of a user seeking information from a document graph; and 4) \textbf{dialog collection} where crowd-collaborators write dialogs based on the generated dialog flows.

\subsection{Document Collection}
For document collection, we examine several potential domains and select 5 representative ones including public services, technology, insurance, health care services, and wikiHow.  For each domain, documents are selected based on two criteria: 1) the documents should be rich in structural types; 2) each document should have links to other documents so that we can test the ability of machines to reason over multiple documents. We design a simple ranking score based on these criteria and select the top-ranked documents for each domain. 



\subsection{Document Graph Construction}

Documents from different domains or sources have vastly different formats (HTML, PDF, etc). Towards building scalable dialog systems across domains, it is important to have a unified format for encoding heterogeneous semantic structures in documents. Bear in mind that our target is to preserve those structures in the document context. This is unlike knowledge graphs and event graphs \cite{fu2020survey,ma2021unstructured,hogan2021knowledge} in which only entities or events are extracted while other context information is discarded.



\begin{figure}[htbp]
\begin{subfigure}
    \centering
    \includegraphics[width=0.46\textwidth]{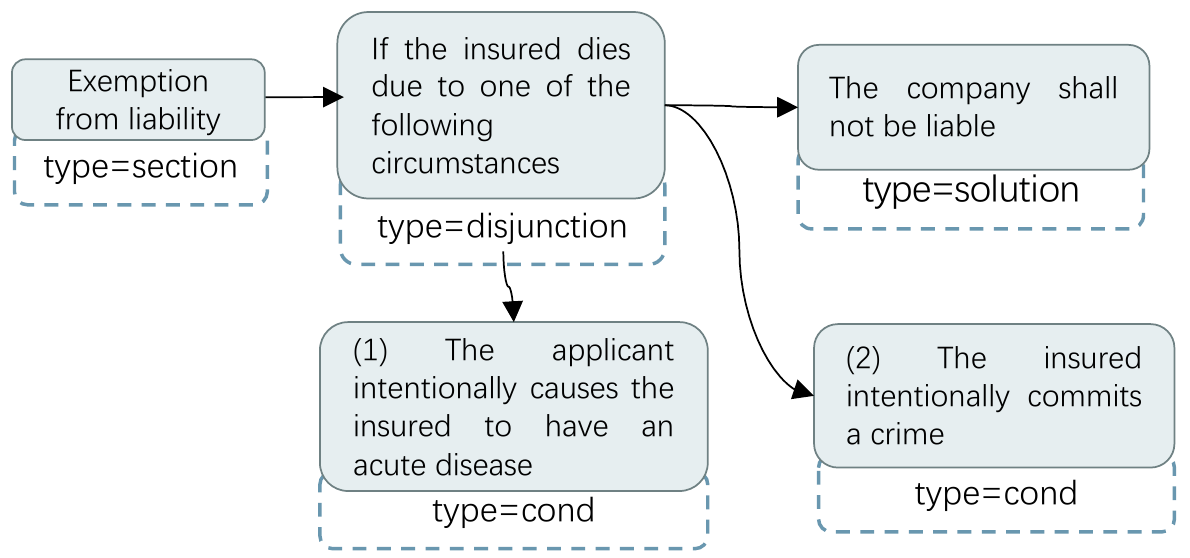}
    \caption{The structure of a disjunction of conditions and the associated solution in the insurance domain.}
    \label{fig:cond-sol}
\end{subfigure}
\vspace{+.5cm}
\begin{subfigure}
    \centering
    \includegraphics[width=0.48\textwidth]{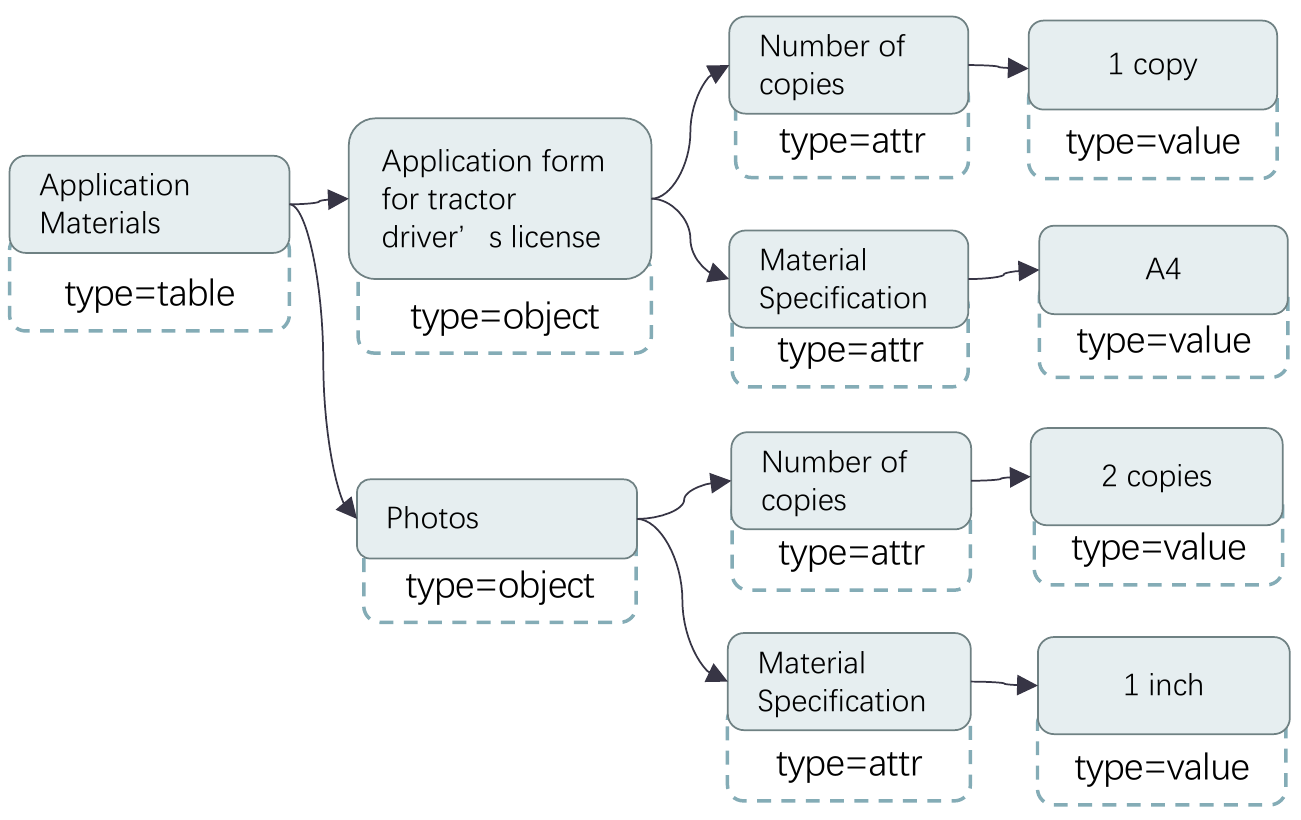}
    \caption{The structure of a table and its objects in the domain of public services.}
    \label{fig:tables}
\end{subfigure}
\end{figure}

\paragraph{Document Graph} is defined as a directed graph where a node corresponds to a span of text in the document. Inspired by property graphs \cite{hogan2021knowledge}, we associate each node with a node type and a set of additional property-value pairs. Each domain has a root node that connects to domain documents via title hierarchy. 

A number of node types are defined to cover common discourse relations exhibited in multiple domains \cite{das2018constructing, stede2019connective}. These include \textbf{section} type to denote section titles in documents.  The types of \textbf{disjunction}, \textbf{conjunction}, \textbf{condition}, \textbf{solution}, \textbf{negation} are used to describe the condition-solution relation as depicted in Figure \ref{fig:cond-sol}. The types of \textbf{table}, \textbf{object}, \textbf{attribute}, \textbf{value} are to encode the relations in tables as shown in Figure \ref{fig:tables}. The types of \textbf{sequence}, \textbf{sequence-step} are introduced to indicate the relations of texts in describing procedures such as \texttt{N3} in Figure \ref{fig:example}. Last but not least, the \textbf{see-more} type is used to encode hyperlinks, and the \textbf{ordinary} type is assigned to the nodes belonging to none of the above.

The property-value pairs associated with nodes are used for additional information. For example, each node can be identified with \texttt{docid} and \texttt{nodeid}. Likewise, \textbf{see-more} nodes have properties such as \textit{linked nodeid}. Additionally, we introduce \textit{is-super-leaf} to indicate whether a node should be targeted in the dialog flow generation.

\subsection{Dialog Flow Generation} \label{sec:dialog-flow-gen}
Studies of human behaviors in goal-oriented dialog systems have long recognized the fact that users have hidden agendas \cite{schatzmann-young2009agenda} which direct the interactions between users and chatbots. This is also the idea behind the construction of well-known datasets such as MultiWoz \cite{budzianowski2018multiwoz}. Although the connection between DGDS in information-seeking scenarios and goal-oriented dialog systems has been suggested \cite{feng2020doc2dial, feng2021multidoc2dial}, DGDS have no explicit schemes, thus hindering the agenda-based approach to dialog collection. As an alternative, we exploit the graph structure of the document graph to build up agendas for simulating dialog flows between a user and an agent. Here, a dialog flow is defined as a sequence of goals, each goal corresponds to a node in our document graph. We mark nodes, that can be used as goals, with \textit{is\_super\_leaf} being true using a semi-automatic method.

Our agenda-based procedure for generating a dialog flow is demonstrated in Algorithm \ref{alg:agenda}. Here, the procedure takes as inputs the document graph $G$, the transition probabilities $\xi$, the maximum number of goals $nGoal$, and the initial selected document $d$. The objective is to generate diverse dialog flows based on which crowd contributors can write conversations. For each goal, a prompt can be generated to suggest questions that can be asked about the subtree rooted at the goal node (line 6). For example, given a table in Figure \ref{fig:tables} as a goal, we can generate the corresponding prompt by: (1) randomly selecting some ``objects'' and ``attributes'' as constraints, e.g. \textit{paper size} and \textit{application form}; (2) using templates to convert the constraints to a guideline such as \textit{``write a number of question-answer turns so that the system final answer is A4 - the \textbf{paper size} of the \textbf{application form}''}.

 \begin{algorithm}
\caption{Agenda-based dialog flow generation}\label{alg:agenda}
\begin{algorithmic}[1]
\Require $G$; $\xi=[\xi_{fl},\xi_{inj}, \xi_{outj}]$, $nGoals$, $d$
\Ensure a dialog flow $flow$, that is a list of $goal$, each corresponds to a node in $G$
\State $goal \gets sample\_leaves(G,d.root)$ 
\State $path \gets get\_path(G, d.root,goal)$
\State Push nodes in $path$ to the agenda stack $A$
\While{$len(flow) < nGoals$}
\State $goal\gets$ pop a leaf from A \Comment{last goal}
\State $prompt\gets gen\_prompt(G, goal)$
\State $flow\gets$ append $([goal, prompt])$
\State Sample $act$ based on $\xi$
\If{$act$ is $follow\_up$}
\State $st\gets \text{pop the top from } A$
\EndIf 
\If{$act$ is $in\_jump$}
\State $st \gets \text{sample a random node in} A$ 
\State $\text{pop A till seeing } st$ 
\EndIf
\If{$act$ is $out\_jump$}
\State $d \gets sample\_connected\_doc(G, d)$
\State $st \gets d.root$ 
\EndIf
\State $goal\gets sample\_leaves(G,st.root)$
\State $path \gets get\_path(G, st.root,goal)$
\State Push nodes in $path$ to the agenda stack $A$
\EndWhile
\end{algorithmic}
\end{algorithm}

We use an agenda stack to contain a list of potential goals that a user might switch to (from the last goal). The candidates nearer to the top of the agenda stack are closer to the last goal in the document graph. The action of a user switching from one goal to another is simulated by three factors, the follow-up rate $\xi_{fl}$, the in-jump rate $\xi_{inj}$ and the out-jump rate $\xi_{outj}$. When the action is \textit{follow-up}, users tend to ask about the related information of the recent goal (line 10). If the action is \textit{in-jump}, users ask about some goals further away from the last goal but still close to some goals in the past. The out-jump action, on the other hand, allows us to simulate the situation where users may ask about related documents. The out-jump rate might be increased if the current goal is linked to an outside document via a \textbf{see-more} node. Note that the procedure that samples leaf nodes (line 1, 17) should exclude the visited nodes. 


Our agenda-based flow generation is adaptable to include new types of structures. This is because whenever we need to target a new structural type, we just need to adjust the document graph definition, and design a new prompt generation  while keeping Algorithm \ref{alg:agenda} unchanged. 


\subsection{Dialog Collection}

We ask crowd contributors to write conversations based on our generated dialog flows. We follow the protocol that one writer plays both agent and user roles and completes the whole dialog like \cite{feng2020doc2dial,feng2021multidoc2dial}. To further improve the coherence of the generated dialogs, we ask writers to examine each dialog flow and skip some goals if the goal is not consistent with the rest of the flow.

Once a dialog flow has been double-checked by a writer, he/she is requested to write dialog utterances based on the goals and their associated prompts (see Section \ref{sec:dialog-flow-gen}). Each goal, its prompt, and its context are then presented one by one to a crowd contributor. By context, we mean the path from the graph root node to the goal node and its neighbors. The writer then interchangeably takes the role of a user or an agent with different interfaces (see Appendix \ref{sec:appendix-annotation-platform}). When it is the user's turn, we encourage the writer to pose an under-specified question, which might make the system confused between the goal node and others in the context. When it is the system's turn, the writer is either asked to provide an answer based on the goal node or ask questions to clarify.   Once the system has fulfilled the goal task, the writer should terminate the goal to move to the next one in the flow. Besides utterances, for each turn, the writer needs to provide annotations such as user/system acts, and grounding texts/nodes.

To ensure the quality of the dataset, crowd collaborators were selected and trained for two weeks. After the training period, we sampled several dialogs and provided feedback to writers in a weekly manner. Our task was completed in 3 months and we paid 0.836 RMB per dialog turn.

\section{Data Analysis} \label{sec:data-analysis}

\paragraph{Document Data} Table \ref{tab:documents} lists the number of documents, along with the number of structures by types in {\dsname}. As we can see, documents in the domain of public services are very rich in structural types, whereas wikiHow contains a lot of sequences showing how-to procedures. Although the size of the document collection is still moderate, it is more than 3 times larger than the document collection size in doc2dial and multidoc2dial \cite{feng2020doc2dial,feng2021multidoc2dial}, the previous datasets for DGDS.

\begin{figure}
    \centering
    \includegraphics[width=0.45\textwidth]{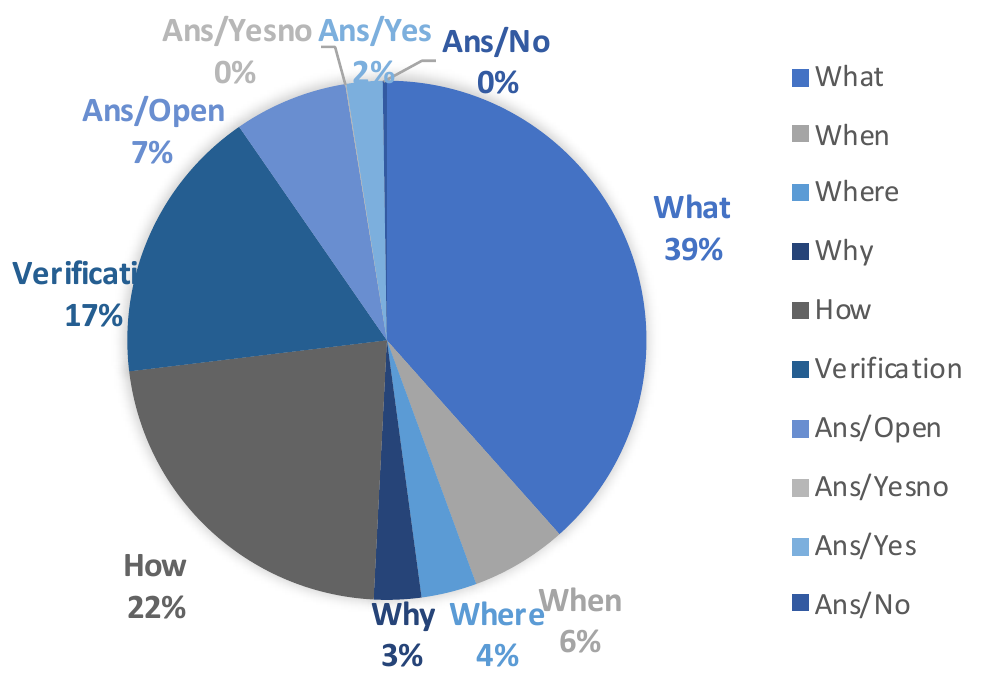}
    \caption{The distribution of user dialog actions.}
    \label{fig:user_acts}
\end{figure}

\begin{figure}
    \centering
    \includegraphics[width=0.48\textwidth]{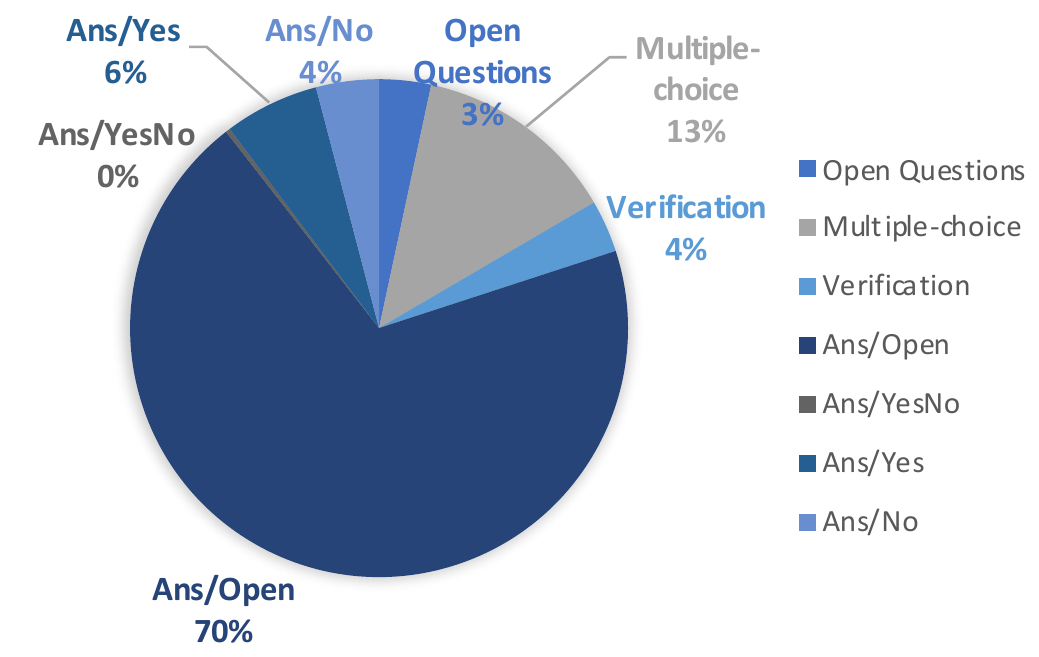}
    \caption{The distribution of system dialog actions.}
    \label{fig:system_acts}
\end{figure}

\paragraph{Dialog Data} {\dsname} contains 6,619 dialogues with 101,994 turns (see Table \ref{tab:dialog-data}). The mean length of user and system utterances are 18.3 and 49.99 words, respectively. Each user utterance is annotated with a dialog state, consisting of a user action (Figure \ref{fig:user_acts}) and some grounding texts (1.39 texts on average). Similarly, each system utterance is annotated with a system action (Figure \ref{fig:system_acts}) and an average number of 1.81 grounding nodes (in the document graph). As we can see from Figure \ref{fig:system_acts}, system questions correspond to about 20\% of the total number of system utterances. This implies that our agenda-based protocol has successfully encouraged crowd-collaborators to include a considerable number of ambiguities.

Table \ref{tab:dialog-data} shows the number of goals for different structural types. Since each goal corresponds to a dialog segment, it can be inferred that although the majority of requests are about plain texts (ordinary), {\dsname} does contain a large number of scenarios grounded on other types of structures.

For evaluation, we define several tasks (see Section \ref{sec:tasks-baselines}) and split the dialog dataset into a training set, consisting of 70\% of dialogues, as well as a validation set of 10\% and a test set of 20\%. 


\begin{table}
    \centering
    \begin{tabular}{p{0.9cm}c|cccc}
        \toprule \textbf{Domain} & \textbf{\#Docs} & \textbf{\#Tab} & \textbf{\#Seq} & \textbf{\#Con} & \textbf{\#Sec} \\
         \midrule Public & 390 & 2747 & 1909 & 644 & 3993\\
         Health  & 403 & 0 & 2 & 22 & 2468 \\
         Insurance & 100 & 0 & 0 & 2423 & 6107 \\
         Wikihow  & 385 & 0 & 658 & 0 & 1900 \\
         Tech.  & 301 & 47 & 771 & 250 & 1526 \\
         \midrule 
         \textbf{All} & \textbf{1579} & \textbf{2794} & \textbf{3340} & \textbf{3339} & \textbf{15994} \\
         \toprule
    \end{tabular}
    \caption{Statistics on the number of documents and the structural types by domain. Here, ``Public'' is short for ``Public Services''. Tab, Seq, Cond, and Sec stand for tables, sequences, conditions/solutions, and sections.}
    \label{tab:documents}
\end{table}

\section{Tasks and Baselines}\label{sec:tasks-baselines}
Inspired by goal-oriented dialog systems, this paper considers three tasks: (1) dialog state tracking; (2) dialog policy learning; (3) response generation. 

\begin{table}
    \centering
    \begin{tabular}{ll|ll}
        \toprule
         \multicolumn{4}{l}{\textbf{Conversation and Goal Statistics}}\\
         \midrule
          \#dialogs & 6619 & \#turns & 101994\\ 
          \#dial (>1 doc) & 2207 & \#doc/dial & 1.42 \\
          \#gr/usr-turn & 1.39 & \#gr/sys-turn & 1.81\\
          usr-turn len & 18.3 & sys-turn len & 49.99 \\
          \midrule
          \#ordinary & 20747 & \#tables & 607\\
          \#sequences & 5707 & \#conditions & 896\\
          \toprule
    \end{tabular}
    \caption{Conversation statistics: the \# of dialogues, turns, the average \# of grounding (\#gr) texts per turn, grounding documents per dialog. Goal statistics: the \# of dialog segments about tables, conditions, etc. }
    \label{tab:dialog-data}
\end{table}

\subsection{Dialog State Tracking}
\label{sec:dst}
Dialog State Tracking (DST) aims at tracking user intentions and key information \cite{dai2021preview,sun2021unsupervised,zhang2022slot}. Specifically, given a dialog history $H$ and the latest user turn, we need to perform (1) user action prediction where the actions are given in Figure \ref{fig:user_acts}, and (2) grounding text matching where the candidate texts are extracted from nodes of the document graph without duplication. 



\paragraph{Baseline Approach} The problem of user act prediction can be formalized as a multi-class classification, where we use RoBERTa \citep{yinhan2019roberta}, BERT \citep{jacob2019bert} and ELECTRA \citep{kevin2020electra} as our baselines. The problem of grounding text matching can also be formalized as a classification problem, where we classify a text as relevant vs irrelevant given the dialog history. Since the number of texts is too large for evaluation, we follow the retrieval then classification approach, where a retrieval model is first used to retrieve candidate texts, and the above models are used to classify if a candidate is relevant or not. 

All in all, BM25 \cite{robertson2009bm25} and Dense Passage Retrieval \cite{karpukhin2020dpr} are used to select a number of candidate texts. And for each of the above classification baselines, we use two independent models for the user action and grounding text prediction.



\paragraph{Evaluation Metrics} For user action prediction, we report micro-F1 (F1) and macro-F1 (ma-F1) which are the mean F1 scores averaged over turns and actions, respectively. For grounding text prediction, micro-F1 is applied to measure the performance of our baselines. In addition, joint accuracy is used to measure the percentage of turns, for which the user action and all the grounding texts are correctly identified. 

\paragraph{Experimental Results} Table \ref{tab:dst} shows that while we can achieve promising results with user action prediction, the task of grounding text matching is still very challenging. For the matching task, 200 candidates are retrieved by BM25 or DPR, and then reranked by the classification models (RoBERTa, BERT, or ELECTRA). It is observable that models based on DPR (e.g., DPR+RT) outperform BM25-based models (e.g. BM25+RT), indicating DPR is a better retrieval method compared to BM25. The best F1 score and joint accuracy, however, are only 57.26\% and 34.55\%, respectively. The low value of F1 on grounding text prediction suggests room for improving both the retrieval and the classification models. The lower value of joint accuracy shows that it is even more challenging to correctly identify all the grounding texts.

\begin{table}
    \centering
    \begin{tabular}{lcccc}
        \toprule \multirow{2}*{\textbf{}} & \multicolumn{2}{c}{\textbf{Act}} & \multicolumn{1}{c}{\textbf{GN}} & Joint\\\cmidrule{2-4}
         &  ma-F1 & F1 &  F1 & Acc\\
         \midrule DPR+RT & \textbf{77.45} & 89.38 &\textbf{57.26} & \textbf{34.55}\\
          BM25+RT & - & - &   38.62 &  15.39\\
        \midrule DPR+BT & 76.95 & \textbf{89.52} &57.15 & 32.22 \\
          BM25+BT & - & - &  42.32 &  19.90\\
        \midrule DPR+ET & 76.06 & 87.88 & 57.02 & 30.29\\
          BM25+ET & - & - &  39.24 & 16.73\\
         \toprule
    \end{tabular}
    \caption{Results on user act prediction (Act) and grounding text prediction (GN) using BM25 and DPR for retrieval. Here, RT is for RoBERTa, BT is for BERT, and ET is for ELECTRA.}
    \label{tab:dst}
\end{table}

\subsection{Dialog Policy Learning}
This task aims at planning the system act and the contents to generate the next response \cite{he2022galaxy}. The input of this task includes (1) the dialog history $H$, (2) the document graph $G$, and (3) the dialog state $DS$; whereas the expected output consists of the system action (Figure \ref{fig:system_acts}) and the grounding nodes set. 

Unlike grounding text matching in DST where we find texts related to user requests, grounding node prediction requires the agent to locate nodes that should be used for system response. For example, when asking about the table in Figure \ref{fig:tables}, users may provide the table name and the attribute names, which are found in the DST task. The dialog policy then infers the ``value'' nodes that contain the answer given the dialog states. By assuming that the dialog states are available, we hypothesize that the agent can fully understand user utterances, thus having a perfect DST module. In practice, this is still a difficult task as seen in the previous section. 


To further simplify the task of grounding node prediction, we formalize it as a classification problem where the agent just needs to predict whether a candidate node should be used or not. Here, a candidate set is selected for each turn by combining the gold system grounding nodes, and 30 distractors, which are chosen randomly from the set constructed by: 1) selecting nodes with texts that match the most recent dialog states; 2) selecting the neighbors (parents, siblings) of the nodes found in (1) as well as those of the gold grounding nodes.

\paragraph{Baseline Approach} For the system act prediction, we also use RoBERTa, BERT and ELECTRA as the classification models where the input is the dialog history, and the output is one of the system act (Figure \ref{fig:system_acts}). For grounding node prediction, we first form the input by concatenating (1) history: the two latest utterances in the dialog history; (2) the dialog state; (3) the structure information obtained by sequentializing the path leading to the candidate node in the document graph \cite{wan2021does}; and (4) the candidate node. Before each segment (1-4) of the input, we add a special prompt to indicate its semantics. We then use these classification models to predict the relevance of the candidate node.

For the ablation study, we consider two variants for each of these baselines for the grounding node prediction. The first one treats a document as a sequence of texts and replaces the structure information with the context window of the node in the original document, we refer to this as (- structure). The second variant excludes the dialog state information, which is referred to as (- dialog state). 


\paragraph{Experimental Results} The same metrics, which are used to evaluate dialog state tracking, are used here for evaluating dialog policy models. The experimental results are shown in Table \ref{tab:dap}, from which several observations can be drawn. Firstly, the best baseline can only achieve macro-F1 of 46.66\% on system action prediction, showing that this task is more difficult than user action prediction. The main reason for the difficulty of this task is the imbalance in the action distribution (see Figure \ref{fig:system_acts}), an issue that requires further attention. Secondly,  both document structures and dialog states are essential for grounding node prediction, since the performance drops significantly without either one of them. And finally, despite having the full information of dialog states and a simplified formalization with only 30 most potential distractors, the best performance we can obtain is only 43.13\% joint accuracy.



\begin{table}
    \centering
    \begin{tabular}{p{1.5cm}cccc}
        \toprule \multirow{2}*{\textbf{Model}} & \multicolumn{2}{c}{\textbf{Act}} & \multicolumn{1}{c}{\textbf{GN}} & Joint\\\cmidrule{2-4}
         &  ma-F1 & F1 &  F1 & Acc\\
         \midrule RoBERTa & \textbf{46.66} & \textbf{82.87} & \textbf{81.67} & \textbf{43.13}\\
          -structure & - & - &  77.18 & 35.51\\
          -states & - & - &  73.71 & 32.49\\
         \midrule BERT & 45.00 & 82.58 &  81.40 & 41.75\\
          -structure & - & - &  75.61 & 32.08\\
          -states & - & - &  73.14 & 31.21\\
         \midrule ELECTRA & 42.77 & 81.91 &  81.31 & 40.31\\
          -structure & - & - &  75.66 & 32.18\\
          -states & - & - &  73.17 & 30.53\\
         \toprule
    \end{tabular}
    \caption{Results on system action prediction (Act) and grounding node prediction (GN).}
    \label{tab:dap}
\end{table}

\subsection{Response Generation}
This task aims at generating a natural language response based on the given system act and grounding nodes set. The response can be a clarifying question or an answer. The input includes (1) dialog history $H$, (2) system act $a_s$, and (3) grounding nodes set $\mathbb{N}_g$. The target output is a system response $r$ consistent with the chosen action, the planned contents, and the history.

\paragraph{Baseline Approach} We use three encoder-decoder generative models, Pegasus \citep{zhang2020pegasus}, BART \citep{lewis2020bart} and T5 \citep{colin2020t5}, as baselines. Here, the encoder takes the concatenated sequence of all information as inputs, and the decoder generates the response. 

To study the impact of the system action prediction task, we consider the variants of three baselines where the system actions are not included as input.


\paragraph{Experimental Results} We use BLEU\footnote{\href{https://github.com/nltk/nltk}{https://github.com/nltk/nltk}} \cite{papineni2002bleu}, a commonly used metric to evaluate the performance of the response generation. The experimental results are given in Table \ref{tab:rg}, where two main observations can be found. Firstly, T5 is slightly better than other baselines for response generation. One possible explanation is that T5 has more parameters (see Table \ref{tab:model}) than BART, and the pre-training task of Pegasus is more suitable for the summarization task. It is worth mentioning that the results here are obtained by using gold values of system actions and grounding nodes. In practice, we need to take into account the errors accumulated by the DST model and the dialog policy model. Given that the best joint accuracy of DST and dialog policy are only 34.55\% and 43.13\%, one can see that the performance of response generation is still far from this upper bound. Secondly, while the grounding nodes are undeniably important for response generation, the introduction of system actions has not always been helpful in the previous dataset \cite{feng2020doc2dial}. However, in {\dsname}, we find that having information on system actions can help improve the performance of response generation for all baselines. This partially validates our design choices of system actions in {\dsname}. 

\begin{table}
    \centering
    \begin{tabular}{p{2.3cm}|ccc}
        \toprule \textbf{Model} & \textbf{BLEU-1} & \textbf{BLEU-2} & \textbf{BLEU-4} \\
        \midrule Pegasus & 59.85 & 53.48 & 44.56 \\
        $\text{Pegasus}_\texttt{w/o act}$ & 58.12 & 52.09 & 43.59 \\
        \midrule BART & 60.35 & 54.00 & 45.03 \\
        $\text{BART}_\texttt{w/o act}$ & 58.75 & 52.62 & 44.06 \\
        \hline T5 & \textbf{60.97} & \textbf{54.41} & \textbf{45.20} \\
        $\text{T5}_\texttt{w/o act}$ & 59.30 & 52.98 & 44.19 \\
        \toprule
    \end{tabular}
    \caption{Response generation results for Pegasus, BART and T5. $\texttt{w/o act}$ means the system actions are not included as input.}
    \label{tab:rg}
\end{table}



\section{Conclusion}

This paper presented {\dsname}, a novel dataset for DGDS for information seeking. Unlike prior datasets, {\dsname} contains examples that simultaneously test the ability of machines to comprehend heterogeneous documents and clarify user information needs. We proposed three main tasks associated with {\dsname}: (1) dialog state tracking, which tracks user intentions during the conversation, (2) dialog policy learning, which plans the next system action and contents, and (3) response generation, which is to generate system responses based on the outputs of the dialog policy. We then presented baselines for our tasks using several contemporary models. Our experimental results showed that: 1) Both dialog states information and document structure information are important for the task of dialog policy learning; 2) Planning system actions helps improve response generation; and 3) The tasks of dialog state tracking and dialog policy learning, which are essential for response generation, are still very challenging with joint accuracy of only 34.55\% and 43.13\% respectively. We hope that our dataset and such observations will be helpful for future research in this direction.


\section*{Limitations}

Towards a practical document-grounded dialog system, some problems have not been addressed in this work. (1) The task of automatic construction of document graphs deserves further attention. Although there exist automatic solutions for parsing discourse relations, the results are still far from desirable for Chinese. As a result, manual post-processing was required for the construction of document graphs, which can be daunting, particularly for domains with many documents. (2) {\dsname}  has yet to include unanswerable cases. Although such samples can be created with a data recomposition step, due to time limitations, we have to leave this consideration for the future versions of {\dsname}. 



\section*{Ethics Statement}
Several ethical issues need our attention. Firstly, more research should be done to make sure the robustness and effectiveness of document-grounded dialog systems. Without careful consideration, such systems will inconvenience both users and the organizations that own the dialog systems. Secondly, although we can control the diversity of our dataset by adjusting the sampling ratios, the conversations might still contain some level of biases, for which more careful examination should be done. Third, our dataset should be used only for research purposes. For Health domain, the responses must not be taken for diagnosis. Finally, although our collected dialogs contain no privacy sensitive data, a part of the documents has usage constraints, and thus we can only publish part of our dataset. The full dataset can be shared upon usage agreements, and for research purposes only.

\section*{Acknowledgements}
We would like to thank the anonymous reviewers for their insightful comments. We also like to thank Dr. Song Feng and Dr. Bowen Yu for their helpful suggestions. This work was supported by Alibaba Innovative Research project ``Document Grounded Dialogue System''.

\bibliography{agendas}
\bibliographystyle{acl_natbib}

\appendix
\section{Dialog Collection Process} \label{sec:appendix-dialog-collection}

Each dialog flow is presented to crowd-collaborators with prompts that suggest questions related to selected goals in the dialog flow. We design different prompts associated with different types of nodes:

\begin{itemize}
    \item Asking about Tables :We assume that all tables contain a four-level structure \textit{table-object-attribute-value}. We consider each table as a list of key objects and some attributes with them. Inspired by \cite{pasupat-liang2015compositional, chen2020hybridqa}, we design three types of prompt for tables: 1. asking for the general information of a table, such as \textit{``what material do I need to offer?''} 2. asking for the general information of an object in the table, such as \textit{``can you tell me more about the first material?''} 3. asking for a value of the object attribute, such as \textit{``How many copies of the first material do I need to bring?''}. 
    \item Asking about Sequence: Similar to tables, we design three patterns for asking about the sequence: 1. asking for the general information of a sequence, such as \textit{``what should I do for the application?''} 2. asking for the general information of a step in the sequence, such as \textit{``can you tell me more about step one?''} 3. asking for specific information for one step, such as \textit{``how long does step one take?''}
    \item Asking about Condition/Solution: Depending on the final system answer, we design three QA patterns for asking about conditions. We will randomly select a pattern from YES/NO/CONDITIONAL/SOLUTION as a prompt. For the first two patterns, the user need to ask a verifying question like \textit{``can I apply for this fund?''}, and the final answer for system must be YES or NO after checking some conditions. For CONDITIONAL, the user needs to ask a question to know the conditions he needs to meet, such as \textit{``I want to apply this fund, what do I need?''}. For SOLUTION, the user will explicitly say some conditions, and the final system answer should be the solution for that. For example, the user may ask \textit{``I am a 35 year old worker, which insurance can I apply for?''}.
    \item Asking about Ordinary: Ordinary nodes correspond to unstructured texts where we would like to include samples similar to the task of machine reading comprehension. As a result, we ask crowd collaborators  to make up questions based on the node text, so that a span in the node text can be used to answer the question.
\end{itemize}

\begin{figure*}[htbp]
\centering
\begin{subfigure}
    \centering
    \includegraphics[width=0.90\textwidth]{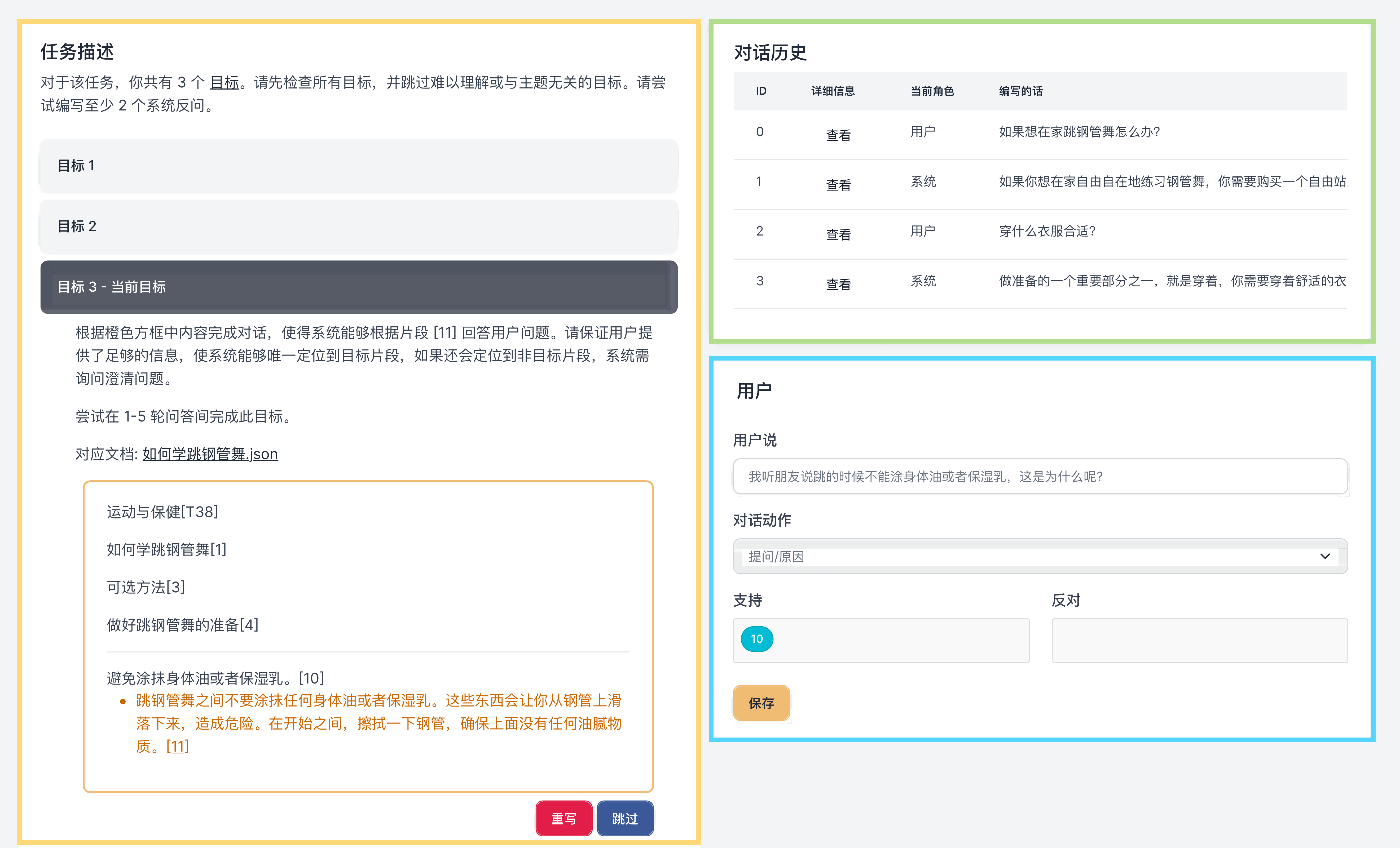}
    \caption{Annotation interface to write a user sentence. The dialog flow we generated is shown in the yellow box, and the dialog history is shown in the green box. The crowdsourcers are asked to write a sentence, select dialog act and grounding nodes in the blue box, to lead the dialog to the orange target text of the current goal.}
    \label{fig:user-inter}
\end{subfigure}
\vspace{+.5cm}
\begin{subfigure}
    \centering
    \includegraphics[width=0.90\textwidth]{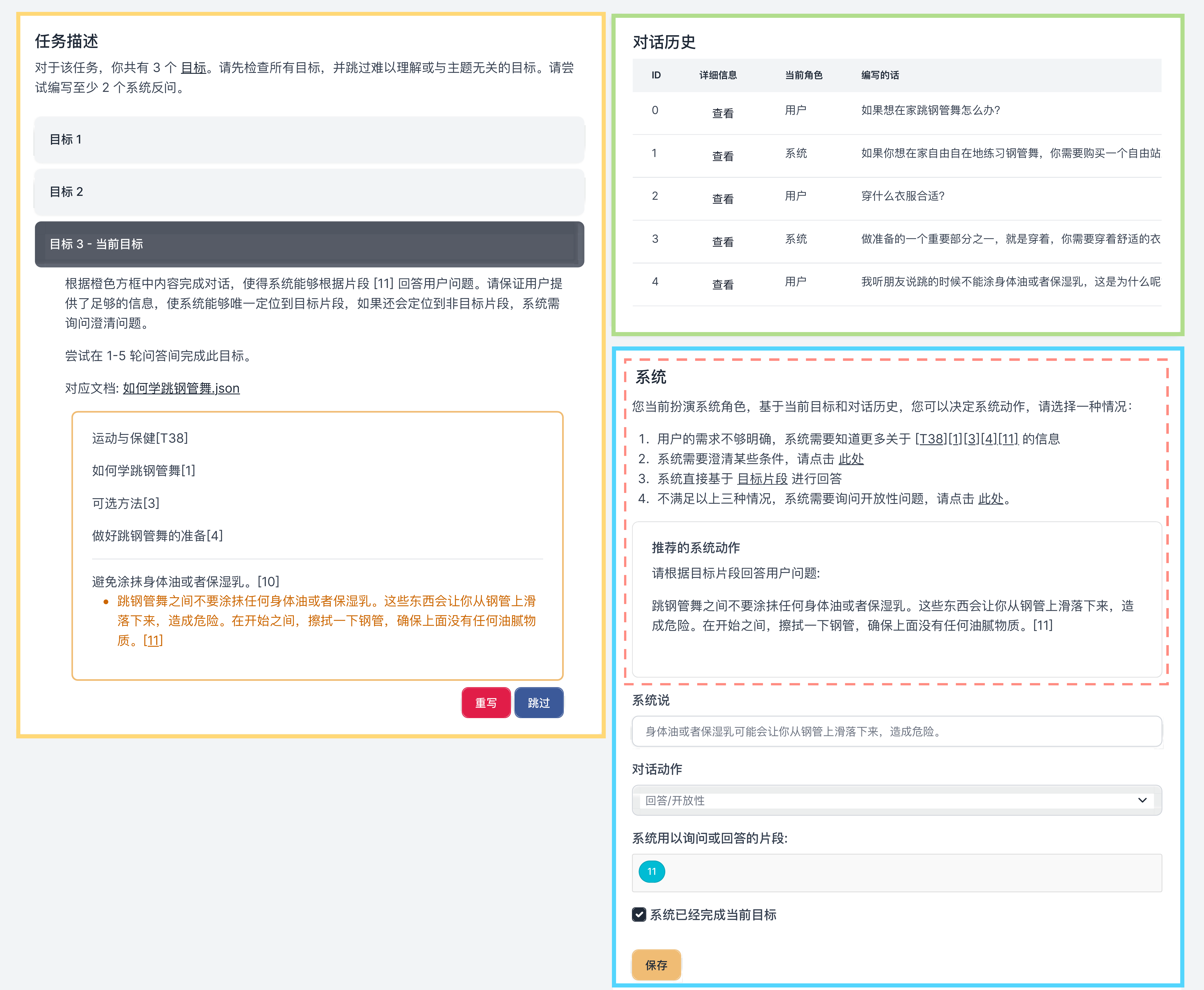}
    \caption{Annotation interface to write a system sentence. Compared to the interface of user turn, we will have more suggestions to help crowdsourcers make up an utterance, which is shown in the red box.}
    \label{fig:sys-interface}
\end{subfigure}
\end{figure*}

The writer is asked to think about a consistent information-seeking situation based on the flow and has the option to skip some goals in the dialog flow. The writer then interchangeably takes the role of a user or an agent with different interfaces. When it is the user's turn (see Figure \ref{fig:user-inter}), we encourage the writer to pose an under-specified question, which might make the system confused between the goal node and others in the context. When it is the system turn (see Figure \ref{fig:sys-interface}), the writer is either asked to provide an answer based on the goal node or ask questions to clarify. For example, a writer can exploit the section nodes to write questions so that the next user's answer can help the agent to better target the goal node in the document graph. Likewise, a writer can select condition nodes (see Figure \ref{fig:cond-sol}) to write questions so that the user answer can help to answer the user question about the condition/solution structure.  Once the system has fulfilled the goal task, the writer should terminate the goal to move to the next one in the flow. Besides utterances, for each (user or system turn) the writer needs to provide annotations such as (user/system) acts, node texts that entail user utterances, and grounding nodes for system questions/answers.
\label{sec:appendix-annotation-platform}

\section{Experiments}
\label{sec:appendix-experiments}
\begin{table*}
    \centering
    \begin{tabular}{c|cccc}
        \toprule \textbf{Model} & \textbf{Variant} & \textbf{Parameters} & \textbf{HuggingFace Pre-trained Model} & \textbf{Corpus}\\
         \midrule BERT & base & 102M & \small\tt{hfl/chinese-bert-wwm-ext} & - \\
         RoBERTa & base & 102M & \small\tt{hfl/chinese-roberta-wwm-ext} & - \\
         ELECTRA & base & 102M & \small\tt{hfl/chinese-electra-base-discriminator} & - \\
         \midrule Pegasus & base & 214M & \small\tt{uer/pegasus-base-chinese-cluecorpussmall} & 14G\\
         BART & base & 116M & \small\tt{fnlp/bart-base-chinese} & 200G\\
         T5 & base & 215M & \small\tt{uer/t5-base-chinese-cluecorpussmall} & 14G\\
         \toprule
    \end{tabular}
    \caption{The detail of the pre-trained model for all baselines. Exact corpus size of the RoBERTa model is not reported by the owner.}
    \label{tab:model}
\end{table*}

\begin{table*}
    \centering
    \begin{tabular}{c|p{2.4cm}ccccc}
        \toprule \multirow{2}*{\textbf{Task}} & \multirow{2}*{\textbf{Model}} & \multirow{2}*{\textbf{Best Epoch}} & \multicolumn{2}{c}{\textbf{Runtime}} & \multicolumn{2}{c}{\textbf{Score}} \\\cmidrule{4-5} \cmidrule{6-7} & &  &\textbf{Training} & \textbf{Inference } & \textbf{Validation} & \textbf{Test}\\
         \midrule 
         \multirow{10}*{DST} & DPR & 47 / 50 & 21.3h & 0.1h & \textbf{75.06} & 73.81\\
         & BERT$_{act}$ & 3 / 5 & 1.6h & < 0.1h & \textbf{89.91} & 89.52 \\
         & BERT$_{bm25}$ & 5 / 5 & 17.3h & 4.3h & \textbf{89.69} & 89.39 \\
         & BERT$_{dpr}$ & 3 / 5 & 16.8h & 3.3h & \textbf{84.52} & 83.67 \\
         & RoBERTa$_{act}$ & 5 / 5 & 1.7h & < 0.1h & \textbf{89.73} & 89.38 \\
         & RoBERTa$_{bm25}$ & 4 / 5 & 19.4h & 4.2h & \textbf{90.32} & 89.95 \\
         & RoBERTa$_{dpr}$ & 2 / 5 & 18.4h & 2.0h & \textbf{84.78} & 83.99 \\
         & ELECTRA$_{act}$ & 5 / 5 & 1.6h & < 0.1h & 87.34 & \textbf{87.88} \\
         & ELECTRA$_{bm25}$ & 4 / 5 & 17.4h & 3.9h & \textbf{90.50} & 90.09 \\
         & ELECTRA$_{dpr}$ & 3 / 5 & 16.6h & 3.5h & \textbf{85.29} & 84.31 \\
         \midrule
         \multirow{12}*{DPL} 
         & BERT$_{act}$ & 2 / 5 & 1.7h & < 0.1h & 81.61 & \textbf{82.58} \\
         & BERT$_{full}$ & 5 / 5 & 25.0h & 0.5h & \textbf{89.16} & 81.40 \\
         & - structure & 5 / 5 & 26.6h & 0.6h & \textbf{82.75} & 75.61\\
         & - dialog state & 5 / 5 & 16.5h & 0.3h & \textbf{84.48} & 73.14 \\
         & RoBERTa$_{act}$ & 5 / 5 & 1.5h & < 0.1h & 82.38 & \textbf{82.87} \\
         & RoBERTa$_{full}$ & 5 / 5 & 24.3h & 0.5h & \textbf{89.67} & 81.65 \\
         & - structure & 5 / 5 & 30.8h & 0.6h & \textbf{84.44} & 77.18\\
         & - dialog state & 5 / 5 & 18.7h & 0.3h & \textbf{84.86} & 73.71 \\
         & ELECTRA$_{act}$ & 5  / 5 & 1.7h & < 0.1h & \textbf{81.96} & 81.91 \\
         & ELECTRA$_{full}$ & 5 / 5 & 22.5h & 0.5h & \textbf{88.28} & 81.31 \\
         & - structure & 5 / 5 & 26.6h & 0.6h & \textbf{82.24} & 75.66\\
         & - dialog state & 5 / 5 & 15.7h & 0.3h & \textbf{83.63} & 73.17 \\
        \midrule
        \multirow{6}*{RG}
         & Pegasus & 18 / 20 & 13.0h & 1.4h & 52.39 & \textbf{52.63} \\
         & Pegasus$_\texttt{w/o act}$ & 18 / 20 & 12.6h & 1.4h & \textbf{51.41} & 51.27 \\
         & BART & 19 / 20 & 10.7h & 0.8h & 53.04 & \textbf{53.13} \\
         & BART$_\texttt{w/o act}$ & 16 / 20 & 13.1h & 0.8h & 51.72 & \textbf{51.81} \\
         & T5 & 19 / 20 & 12.0h & 1.2h & 53.18 & \textbf{53.53} \\
         & T5$_\texttt{w/o act}$ & 20 / 20 & 13.8h & 1.2h & 51.77 & \textbf{52.16} \\
         \toprule
    \end{tabular}
    \caption{The average runtime and validation performance for all baseline models. For the validation score, we use a sample-level F1 score for the dialog state tracking and dialog policy learning baselines. For response generation, we use the average value of BLEU-\{1, 2, 4\}. We also give the sample-level score of the test set, as a comparison of the validation set results.}
    \label{tab:efficient}
\end{table*}

The implementation is in PyTorch and the pre-trained models we used are from HuggingFace Transformers\footnote{\href{https://huggingface.co}{https://huggingface.co}}. The information of them is shown in the Table \ref{tab:model}. For dialog act prediction, we use an MLP to map the pooling output of the pre-trained models to the action space. For all experiments, we evaluate the model at the end of each epoch and select the best-performed checkpoint. But for response generation baselines we only evaluate the last 5 epochs, since the inference stage is time-consuming. AdamW is used to optimize the parameters, with 1e-08 epsilon and 0.01 weight decay. All experiments are performed on one Tesla V100 with 32GB memory. The average runtime of training and inference for each baseline and best validation performance are shown in Table \ref{tab:efficient}.

\paragraph{Hyperparameters for DST baselines}
\label{sec:DST hyperparameters}
For user act prediction, we fine-tune the baselines for 5 epochs with a batch size of 8. We use a learning rate of 5e-6, linear scheduling without warmup. For the DPR, we implement it by ourselves, using a Chinese version of pre-trained RoBERTa as the backbone network. We train the model for 50 epochs with a batch size of 12. The learning rate is 3e-6, linear scheduling with warmup steps of 500. We use one BM25 negative passage per query in addition to in-batch negatives. And the FAISS\footnote{\href{https://github.com/facebookresearch/faiss}{https://github.com/facebookresearch/faiss}} is used to speed up the vector search. For the grounding text prediction baselines, we follow the SC\footnote{\href{https://huggingface.co/docs/transformers/model\_doc/bert\#transformers.BertForSequenceClassification}{https://huggingface.co/docs/transformers/model\_doc/\\bert\#transformers.BertForSequenceClassification}} example from HuggingFace \cite{wolf2020transformers}. We use the top-20 retrieved samples, except for the gold ones, as negative samples for training. We use a learning rate of 2e-5, linear scheduling with warmup steps of 500, for 5 epochs.

\subsection{Hyperparameters for DLP baselines}
For system act prediction, we use the same hyperparameters as the user act prediction in Section \ref{sec:DST hyperparameters}. For baselines and variants in the candidate nodes classification task, we train them for 5 epochs with a batch size of 20. The learning rate is 1e-5 for ELECTRA and 2e-5 for the others, with linear scheduling of 500 warmup steps.

\subsection{Hyperparameters for RG baselines}
We finetune the pre-trained ELECTRA and BART for 20 epochs with a batch size of 12. And the learning rate is 2e-5, also linear scheduling of 500 warmup steps. For the T5 model, we train it for 20 epochs with a batch size of 10. The learning rate is 3e-4 and we use the same linear scheduling as the others. For inference, we set the beam search size as 4, the max generate length as 512, and the length penalty as 1.0. The variants of them use the same hyperparameters since the only difference is the input sequence.

\section{Challenges}
\label{sec:appendix-challenges}
There are different types of challenges exhibited in {\dsname}, which can be divided into three categories: 1) understating documents; 2) understanding dialog context; and 3) abstract reponse generation.


\subsection{Understanding Documents}
\paragraph{Understanding Condition-Solution} the agent needs to be able to recognize fundamental parts of consolution-solution structures and exploit appropriate information to generate system responses.

\begin{figure}
    \centering
    \includegraphics[width=0.48\textwidth]{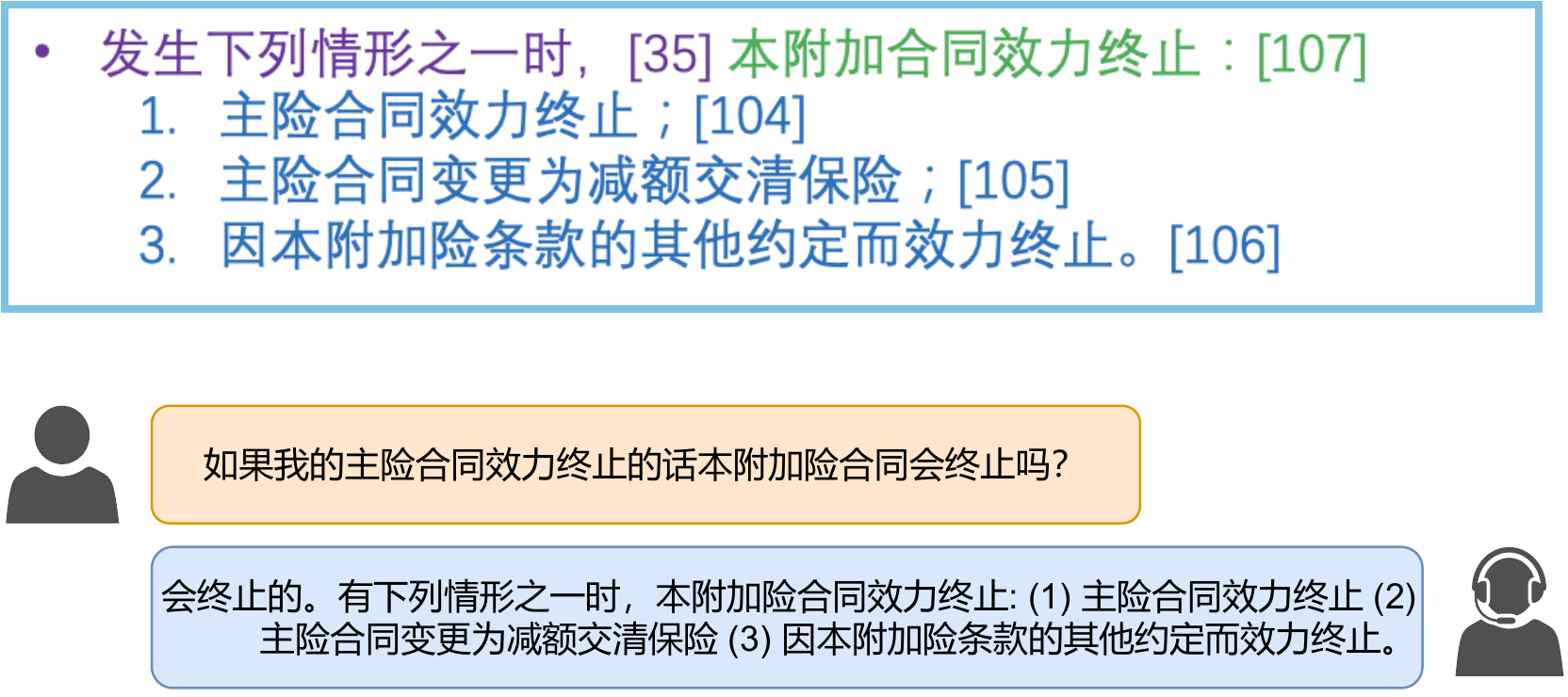}
    \caption{Example of a condition-solution case, where the type of the purple text is condition/disjunction, of the green text is solution and of the blue text is condition.}
    \label{fig:cond-solution}
\end{figure}

\begin{CJK}{UTF8}{gkai}
In Figure \ref{fig:cond-solution}, the user asks \textit{``If the effectiveness of my main insurance contract is terminated, will this additional contract be terminated?''} (如果我的主险合同效力终止的话本附加险合同会终止吗？). Here, the user intention is to know whether he/she meets the conditions of termination of the additional contract. According to the given document, there are three conditions for the termination of the effectiveness of the additional contract. These conditions form a disjunction condition, and the user meets the first condition. Therefore, it is necessary to inform the user that the conditions for termination of this additional contract are met. Here, the agent replies \textit{``Yes, it will be terminated''} (会终止的),  with an additional introduction of all the conditions for the termination of the additional contract.

\paragraph{Understanding Sequence-Step} The agent needs to recognize the sequences and its steps, and know to answer followup questions about specific steps.

\begin{figure}
    \centering
    \includegraphics[width=0.48\textwidth]{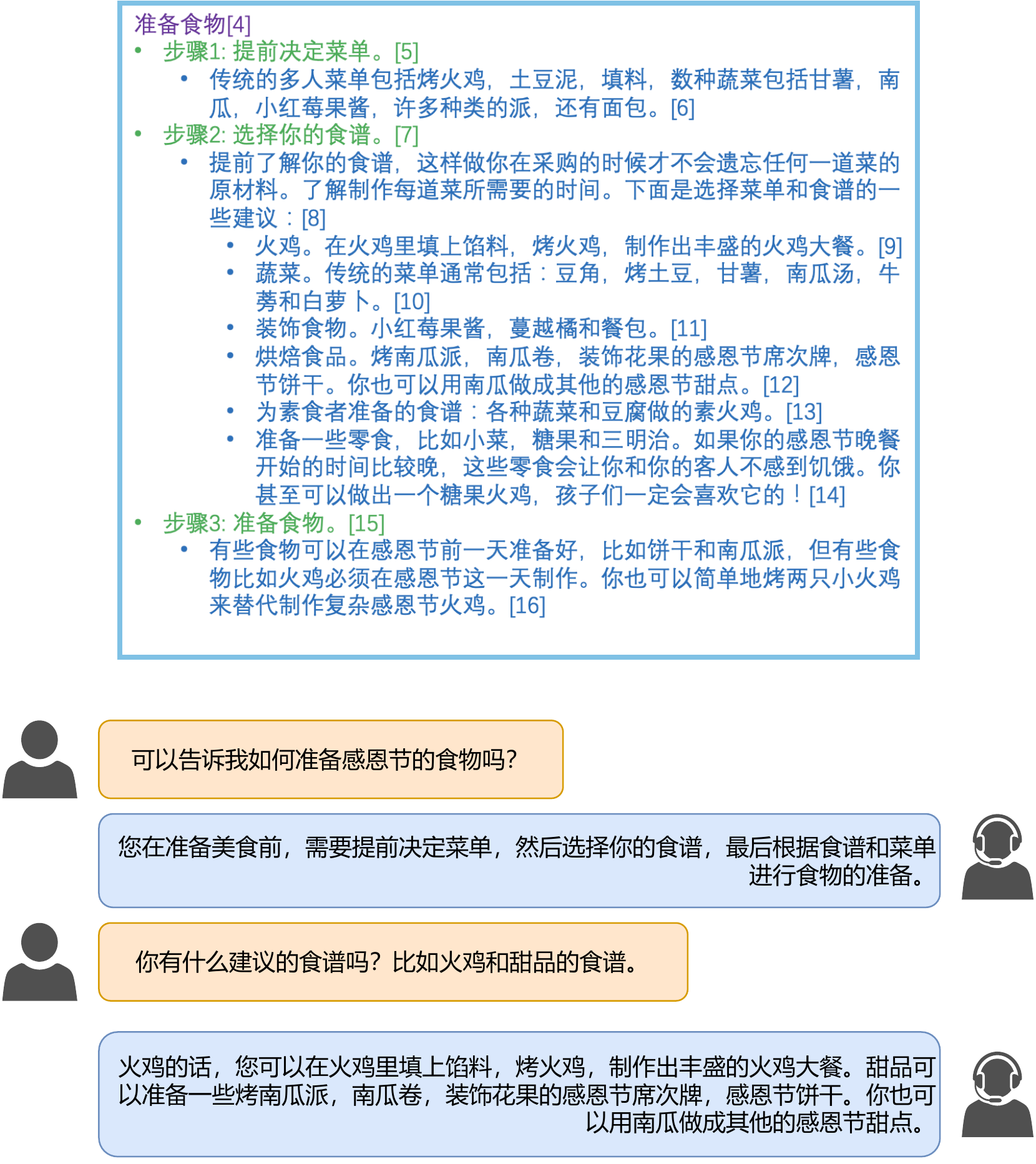}
    \caption{Example of a sequence-step case, where the type of the purple text is sequence, the green text is step and the blue text is ordinary.}
    \label{fig:sequence-step}
\end{figure}

In Figure \ref{fig:sequence-step}, the user asks \textit{``Can you tell me how to prepare Thanksgiving food?''} (可以告诉我如何准备感恩节的食物吗？). The type of ground node is sequence, whose content is \textit{"prepare food"}(准备食物). The agent needs to summarize the steps (the green text) to respond. The user follow-up query is \textit{``Do you have any suggested recipes? Like Turkey and desserts.''} (你有什么建议的食谱吗？比如火鸡和甜品的食谱), targeting on the second step: \textit{``Choose your recipe''} (选择你的食谱). The second step node has two child nodes, node [10] and node [12]. Here, node[10] is \textit{``Turkey. Fill the turkey with stuffing, roast the turkey, and make a rich Turkey meal.''} (火鸡。在火鸡里填上馅料，烤火鸡，制作出丰盛的火鸡大餐。). And node [12] is \textit{``Baked goods. Baked pumpkin pie, pumpkin roll, Thanksgiving dinner plate decorated with flowers, Thanksgiving cookies. You can also make other Thanksgiving desserts out of pumpkins.''} (烘焙食品。烤南瓜派，南瓜卷，装饰花朵的感恩节席次牌，感恩节饼干。你也可以用南瓜做成其他感恩节甜点。). The agent needs to combine the information from the two nodes to answer.

\subsection{Understanding Dialog Context}
\paragraph{User Under-Specified Requests} The user request could be under-specified, and so the agent needs to decide to either ask clarifying questions or provide answer.

\begin{figure}[htbp]
    \centering
    \includegraphics[width=0.48\textwidth]{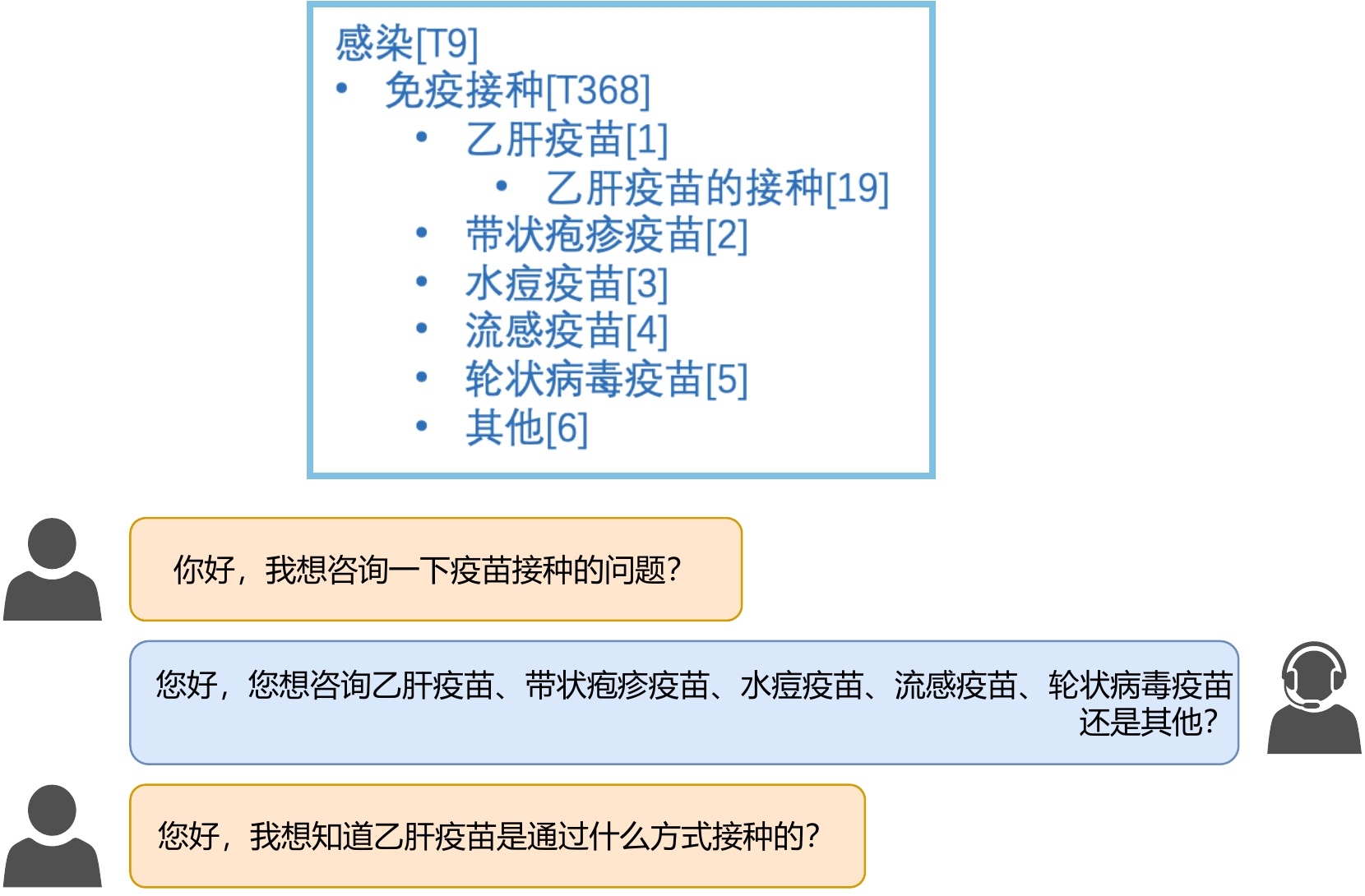}
    \caption{Example of an under-specified request.}
    \label{fig:under-specify}
\end{figure}

In Figure \ref{fig:under-specify}, the user first asks \textit{``Hello, I'd like to ask about vaccination.''} (你好，我想咨询一下疫苗接种的问题。). Since the question is too general, the system asks the user to choose among several choices of vaccines: \textit{"Hello, would you like to consult hepatitis B vaccine, herpes zoster vaccine, chickenpox vaccine, influenza vaccine, rotavirus vaccine or others?"} (您好，您想咨询乙肝疫苗、带状疱疹疫苗、水痘疫苗、流感疫苗、轮状病毒疫苗还是其他？). The user selects one and replies \textit{``Hello, I want to know how the hepatitis B vaccine is vaccinated.''} (您好，我想知道乙肝疫苗是通过什么方式接种的。).

\begin{figure}[htbp]
    \centering
    \includegraphics[width=0.48\textwidth]{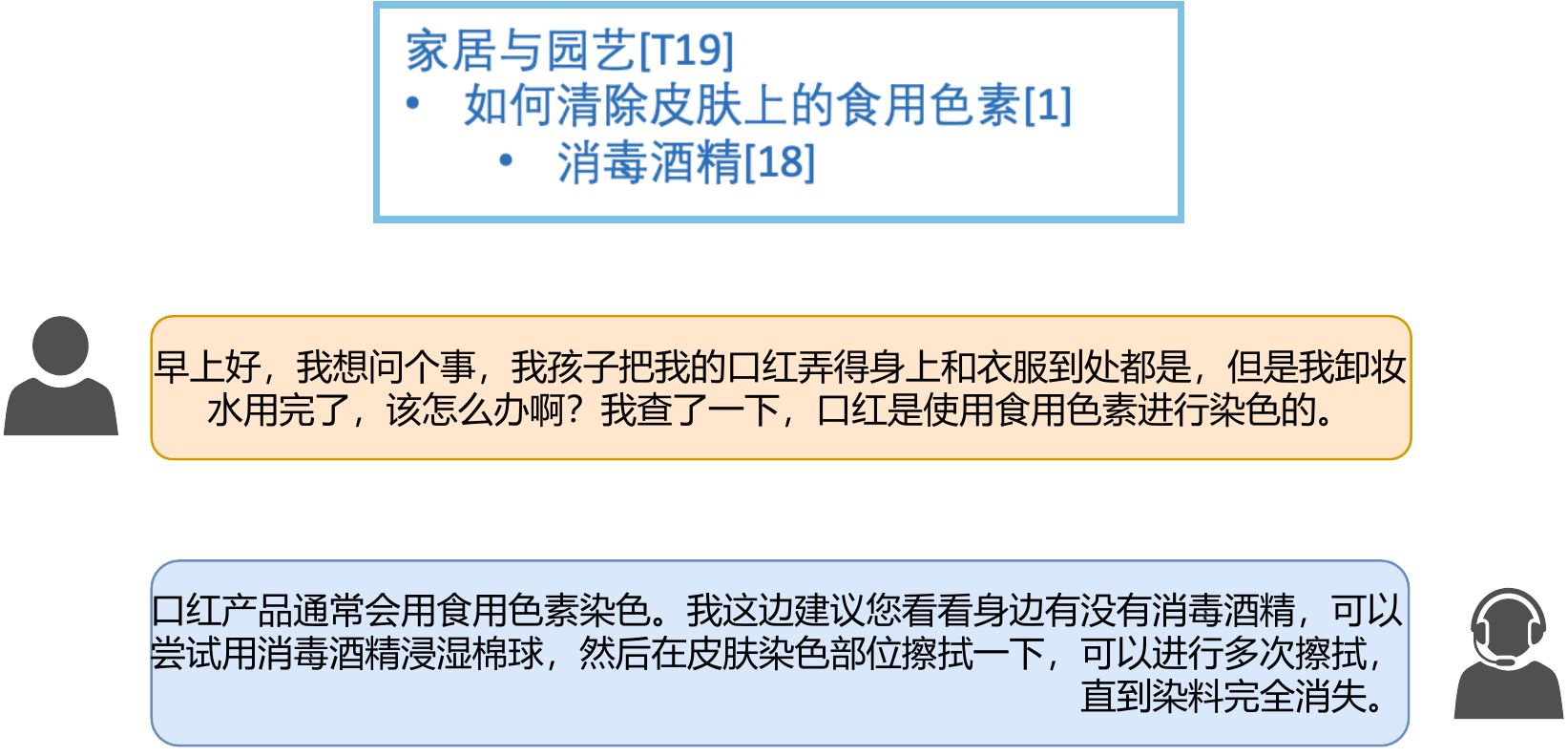}
    \caption{Example of a textual entailment case}
    \label{fig:textual entailment}
\end{figure}

\paragraph{Textual Entailment} The agent needs to be able to recognize if a node text entails the user utterance, thus requiring deeper semantic understanding than text matching. In Figure \ref{fig:textual entailment}, the user asks \textit{``Good morning. I want to ask you something. My child has made my lipstick all over my body and clothes, but I have run out of makeup remover. What should I do? I checked that lipstick is dyed with food coloring.''} (早上好，我想问个事，我孩子把我的口红弄得身上和衣服到处都是，但是我卸妆水用完了，该怎么办啊？我查了一下，口红是使用食用色素进行染色的。). From ``lipstick is everywhere'' and ``I have run out of makeup remover'', we can know that the user intent is to remove the traces of lipstick. In combination with ``I checked that lipstick is dyed with food coloring'', we know that the user request entails \textit{``How to clear the food pigment on the skin''} (如何清除皮肤上的食用色素). 

\paragraph{Coreference} The phenomenon of coreference could occur in a number of user queries. The agent needs to identify these cases and resolve such coreferences for better text understanding.

\begin{figure}[htbp]
    \centering
    \includegraphics[width=0.48\textwidth]{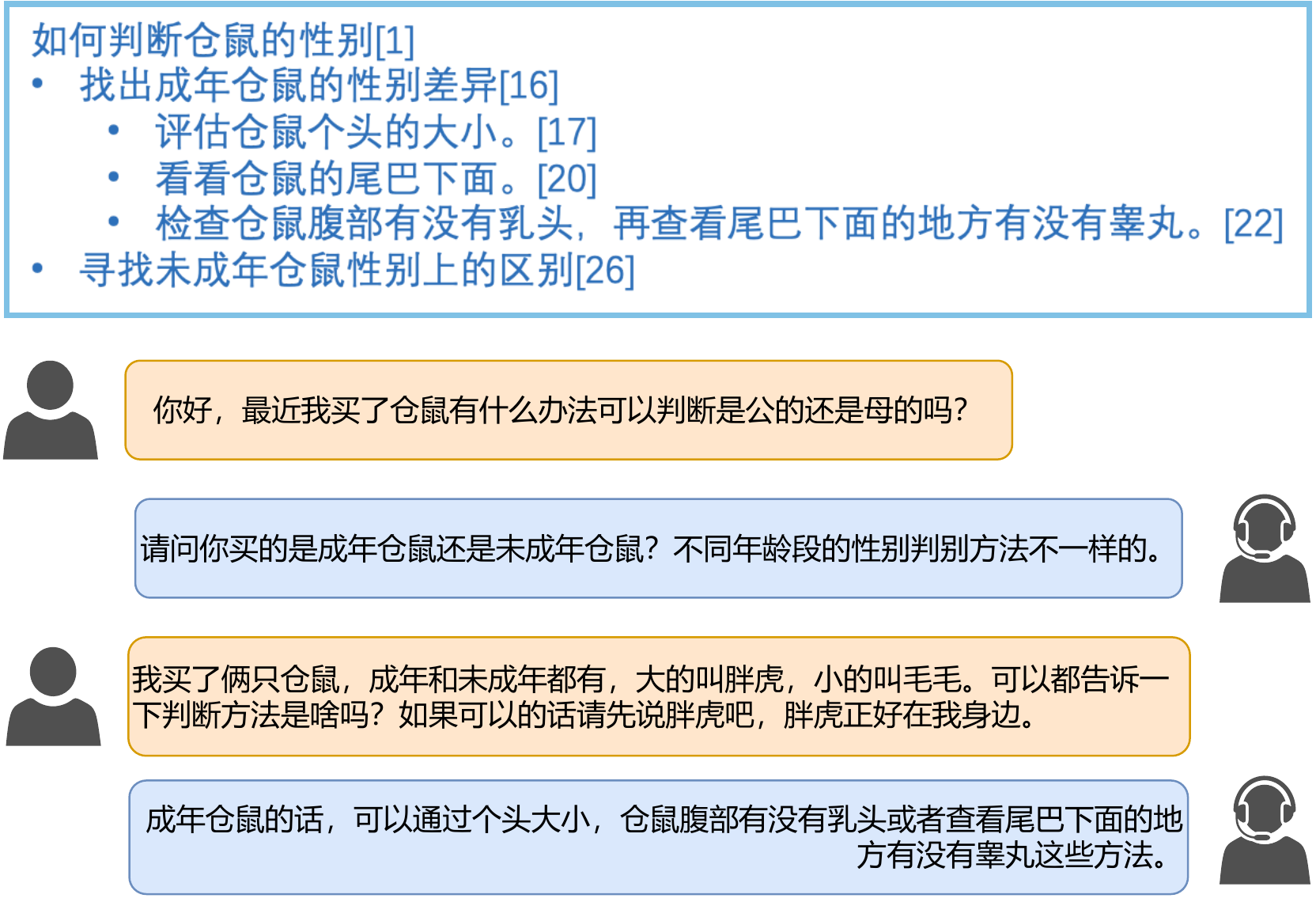}
    \caption{Example of a situation where coreference analysis is needed for understanding user request.}
    \label{fig:coreference}
\end{figure}

In Figure \ref{fig:coreference}, user asks \textit{``Hello, I bought a hamster recently. Is there any way to judge whether it is a male or a female?''} (你好，最近我买了仓鼠有什么办法可以判断是公的还是母的？). Since the current user questions are under specified, the agent guides the user to clarify whether the hamster is an adult. Then the user replies \textit{``I bought two hamsters, an adult and a baby. The big one is Panghu and the small one is Maomao. Can you tell me what the judgment method is? If you can, please tell me Panghu first. Panghu is right beside me.''} (我买了俩只仓鼠，成年和未成年都有，大的叫胖虎，小的叫毛毛。可以都告诉一下判断方法是啥吗？如果可以的话请先说胖虎吧，胖虎正好在我身边。). In this sentence, Panghu and adult refer to the same hamster. And he/she wants to know the answer for the adult hamster first, so the system needs resolve coreference to provide correct answer.

\subsection{Abstractive Response Generation}

\paragraph{Summarization} If the number of grounding nodes for a turn is large, the system needs to be able to summarize the information for more concise responses. 

\begin{figure}[htbp]
    \centering
    \includegraphics[width=0.48\textwidth]{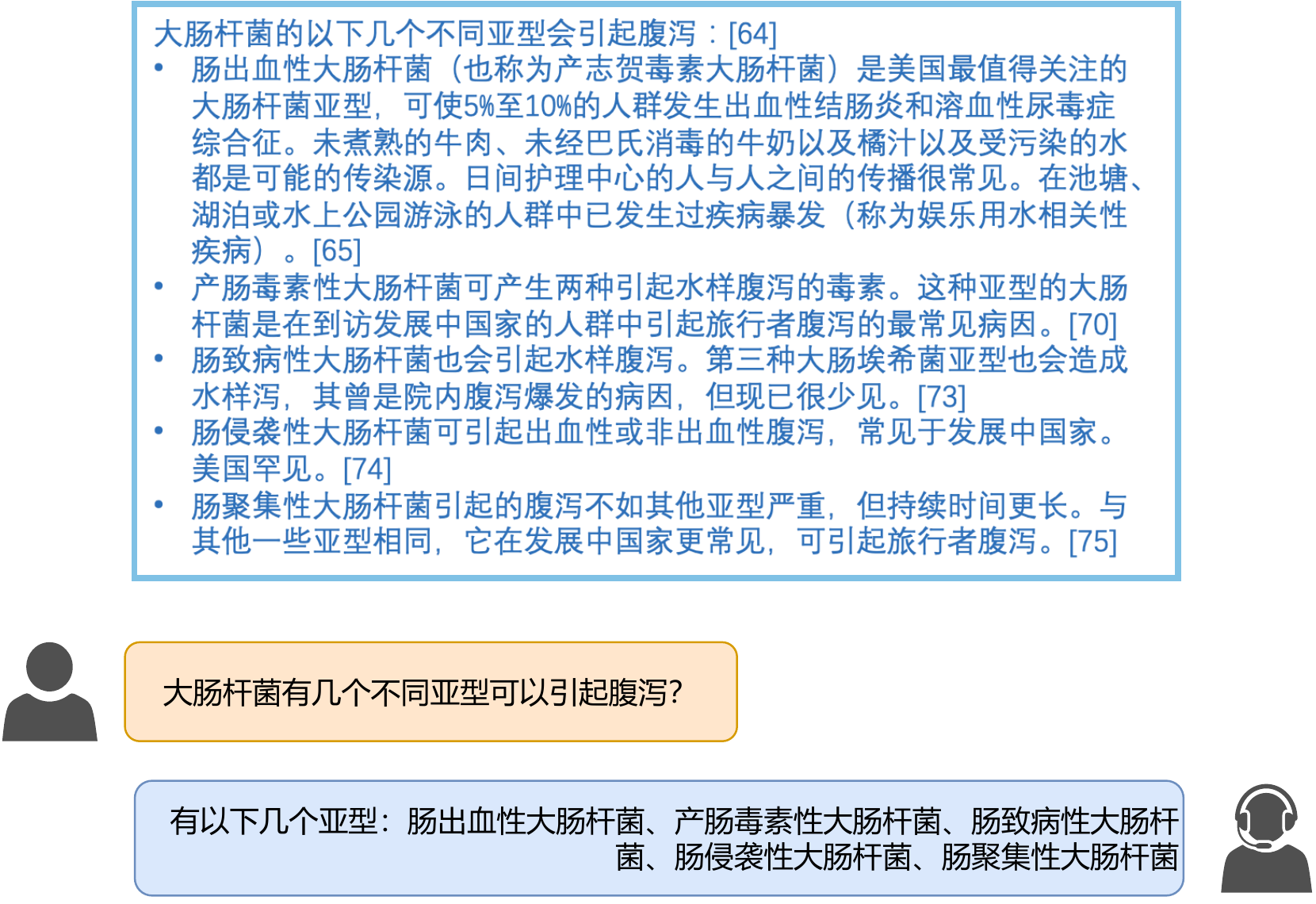}
    \caption{Example of a situation where the agent needs to answer by summarizing the information.}
    \label{fig:summarization}
\end{figure}

In Figure \ref{fig:summarization}, the user wants to know how many different subtypes of E. coli can cause diarrhea (大肠杆菌有几个不同亚型可以引起腹泻？). Since there are 5 grounding nodes, including [65], [70], [73], [74] and [75], the agent needs to summarize information, and give more concise response: \textit{``There are the following subtypes: EHEC, enterotoxigenic Escherichia coli, enteropathogenic Escherichia coli, enteroinvasive Escherichia coli, and enteroaggregative Escherichia coli.''} (有以下几个亚型：肠出血性大肠杆菌、产肠毒素性大肠杆菌、肠致病性大肠杆 菌、肠侵袭性大肠杆菌、肠聚集性大肠杆菌。). 

\paragraph{Common-Sense Knowledge} For some user requests, common-sense knowledge is needed for agents to reason the next system response.

\begin{figure}[htbp]
    \centering
    \includegraphics[width=0.48\textwidth]{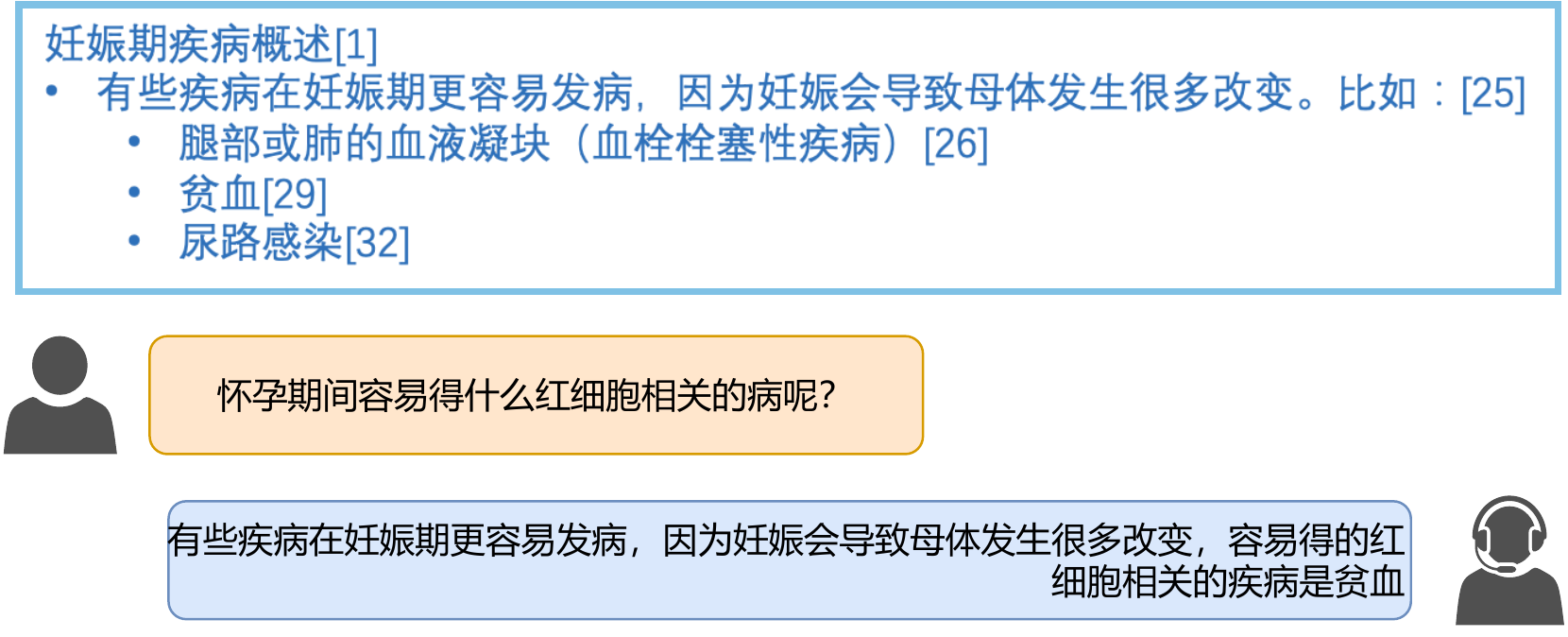}
    \caption{Example of a situation where common-sense knowledge is needed.}
    \label{fig:common sense}
\end{figure}

In Figure \ref{fig:common sense}, the user asks \textit{``What red blood cell related diseases are easy to get during pregnancy?''} (怀孕期间容易得什么红细胞相关的病呢？). In the document, we can identify three diseases that are easy to get during pregnancy：1. Blood clots in the legs or lungs (thromboembolic disease)（腿部或肺的血液凝块（血栓栓塞性疾病)) 2. Anemia（贫血） 3. Urinary tract infection（尿路感染）. The fact that ``Anemia is a red blood cell related disease'' is not in the document, but it is a common knowlwedge. So the response is \textit{``Some diseases are more likely to occur during pregnancy since it will lead to many changes in the mother. The easily acquired red blood cell related disease is anemia.''} (有些疾病在妊娠期更容易发病，因为妊娠会导致母体发生很多改变，容易得的红细胞相关的疾病是贫血。).
\end{CJK}

\end{document}